\documentclass{article}

 \usepackage[preprint]{neurips_2026}

\usepackage[utf8]{inputenc} 
\usepackage[T1]{fontenc}    
\usepackage{hyperref}       
\usepackage{url}            
\usepackage{booktabs}       
\usepackage{amsfonts}       
\usepackage{nicefrac}       
\usepackage{microtype}      
\usepackage{xcolor}         
\usepackage{graphicx}
\usepackage{amsmath}
\usepackage{amssymb}
\usepackage{subcaption}
\usepackage{caption}

\newenvironment{tightlist}
  {\begin{list}{\textbullet}{%
    \leftmargin 0.1in
    \listparindent 0.0in
    \itemindent 0.0in
    \topsep 0in
    \itemsep 0.02in
  }}
  {\end{list}}

\title{Language models struggle with compartmentalization\thanks{Code: \url{https://github.com/vinhowe/compartmentalization}. Eval data: \url{https://doi.org/10.5281/zenodo.20171021}}}

\author{%
  Thomas Vincent Howe \quad David Wingate \\
  Department of Computer Science\\
  Brigham Young University\\
  Provo, UT 84602 \\
  \texttt{th443@byu.edu, wingated@cs.byu.edu}
}

\begin{document}

\maketitle

\begin{abstract}
  
In the training data used by large language models (LLMs), the same latent concept is often presented in multiple distinct ways: the same facts appear in English and Swahili; many functions can be expressed in both Python and Haskell; we can express propositions in both formal and natural language. We show that LLMs can exhibit \textit{compartmentalization}, where they fail to identify and share statistical strength between distinct presentations of unified concepts. In the worst case, LLMs simply learn parallel internal representations of each presentation of the concept, saturating model capacity with redundancies and decreasing sample efficiency with the number of such presentations. \textbf{We also demonstrate that synthetic parallel data can fail to improve this despite being easily learned itself.} Under this framework, we find that, for small models, early multilingual learning is nearly entirely compartmentalized. Finally, all interventions that we study exhibit a phase transition in which their effectiveness depends on the number of distinct presentations, suggesting that the language modeling objective may only inconsistently unify representations.
\end{abstract}

\section{Introduction}

Language models learn from data abundant with distinct presentations of unified concepts. The same facts appear in English and Swahili; many functions can be expressed in both Python and Haskell; we can express propositions in both formal and natural language. We would like the model to recognize these as related, and much of the success of large language models suggests that they often do, at least with enough scale. But language models at frontier scale fail in surprising ways.

\begin{tightlist}
\item In the multilingual setting, \citet{ifergan2024beneathsurfaceconsistency} show that, at 7B scale, language models know more than three times as many facts in any language than the multilingual average, and that task performance across languages does not imply that languages ``learn from each other,'' i.e., share representations for similar tasks.

\item Relatedly, \citet{goldman-etal-2025-eclektic} demonstrate that non-thinking frontier language models consistently fail to recall Wikipedia facts that exist only in a single language in other languages.

\item Within a single language, \citet{calderon2026shelves} find that frontier language models can correctly recite many facts in a pretraining-like Wikipedia context, but not in a QA-like context. In these cases, encoding succeeds but retrieval fails---the model does not functionally ``know'' these facts.

\item In a similar vein, \citet{berglund2024reversalcurse} introduce the Reversal Curse, one of the more well-known and well-replicated frontier failures, wherein LLMs fail to generalize from factual relations of the form "A is B" to the implied but undemonstrated "B is A".
\end{tightlist}

\begin{figure}[t]
  \centering
  \begin{subfigure}[b]{0.329\textwidth}
    \includegraphics[width=\linewidth]{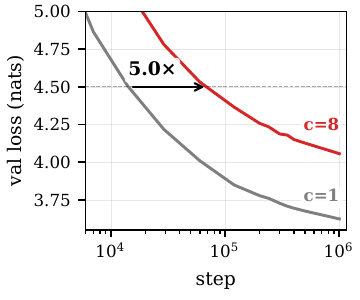}
    \caption{A demonstration of slowdown.}\label{fig:slowdown_def}
  \end{subfigure}
  \hfill
  \begin{subfigure}[b]{0.329\textwidth}
    \includegraphics[width=\linewidth]{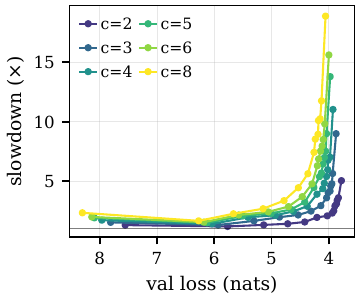}
    \caption{Sample-efficiency cost.}\label{fig:slowdown}
  \end{subfigure}
  \hfill
  \begin{subfigure}[b]{0.329\textwidth}
    \includegraphics[width=\linewidth]{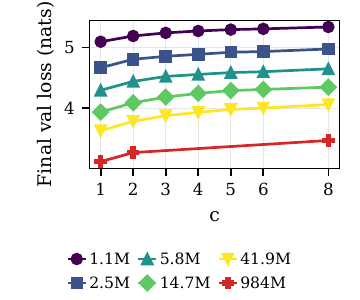}
    \caption{Capacity cost.}\label{fig:plateau}
  \end{subfigure}
  \caption{\textbf{Compartmentalization costs sample efficiency and capacity.}
  \subref{fig:slowdown_def} How slowdown is calculated: at any val-loss target $L$ achieved by some checkpoint of the base $c{=}1$ model (here $L{=}4.5$), find the iterations where $c{=}1$ and $c{=}N$ each cross $L$ and take their ratio. We linearly interpolate to find the intersection point for the compartmentalized model. The example reads $5.0\times$ for $c{=}8$ at $L{=}4.5$ on the 41.9M model.
  \subref{fig:slowdown} Slowdown of each $c$-compartment run vs.\ the $c{=}1$ baseline on the 41.9M model, plotted against the matched val-loss target. Slowdown grows with $c$ across the trajectory.
  \subref{fig:plateau} Final val loss vs.\ $c$, across six model scales (1.1M to 0.98B parameters). Final val loss also grows with $c$.}
  \label{fig:efficiency_capacity}
\end{figure}

In understanding this class of failures, we take inspiration from the field of machine translation, which has long held the interlingua as a target: a shared semantic space that multiple languages share. \citet{johnson-etal-2017-googles} showed that shared encoders can, sometimes, discover representations that generalize across languages they were never jointly trained on. Why then is there so much evidence that frontier models often learn language-specific---and broadly context-specific---representations?

We propose a perspective on this type of failure based on redundancy in representation learning. Specifically, we define \textbf{compartmentalization} as a phenomenon in which a model treats $c$ presentations of a latent data generation process as partially or completely distinct, requiring as many as $c$ distinct representations of that structure. The worst case of compartmentalization is redundancy in the strong sense: the same structure, learned independently, stored as separate representations, with no amortization between them.
We suggest that this has at least two consequences:

\begin{tightlist}
    \item \textbf{Sample inefficiency.} If every latently-similar presentation is treated as a separate prediction task, each presentation benefits only from its own share of the data. In the worst case, where one such presentation accounts for only $\varepsilon$ of the data, it could take $1/\varepsilon$ as many data samples to learn a similar amount of structure under that presentation, assuming no capacity constraints.
    \item \textbf{Capacity competition.} Redundant representations also consume finite representational capacity. Without committing to a particular notion of capacity, we hypothesize that perfect $c$-fold redundancy prevents amortizing across presentations, and therefore requires more capacity to represent than the same data under a single presentation the model represents more cohesively.
    Even with more data, in the capacity-constrained regime, a compartmentalized model might saturate sooner---that is, when evaluated on held-out examples in our worst-case presentation, a model trained on $1/\varepsilon$ as many examples could still plateau above a model the same size trained only on that presentation.
\end{tightlist}

Our contributions within this framework are six-fold.

\section{Contributions}

\begin{tightlist}
\item We define a data augmentation we show allows us to induce compartmentalization in language models, and demonstrate that within this construction language models pay sample-efficiency and capacity costs that scale with $c$ across six model sizes from 1.1M to $\sim$1B parameters.
\item We provide existence proofs that these costs are not fundamental, with two separate methods that each provide unified solutions SGD on its own does not find: models that can predict within compartments without a compartmentalization tax.
\item We find that paired ``translation'' data, despite being easily learned by the model, fails to reduce compartmentalization until a $c$-dependent phase transition that we replicate at 1B parameter scale---we demonstrate that this is accelerated to lower $c$ by weight decay.
\item We show that an auxiliary contrastive representation alignment objective exhibits a similar $c$-dependence; for $c{=}2$ it provides no improvement in 1M steps, but improvement grows with $c$.
\item We propose an operational measure of compartmentalization for models trained on existing datasets with multiple distinct presentations and apply it to multilingual training.
\item With models trained on a biography/Q\&A dataset, we observe destructive capacity competition between multiple formats encoding the same data, with a compartmentalized model as a baseline.
\end{tightlist}

\section{A worst-case model of compartmentalization}
\label{sec:worst-case}

\begin{figure}
  \centering
  \begin{subfigure}[b]{0.49\textwidth}
    \includegraphics[width=\linewidth]{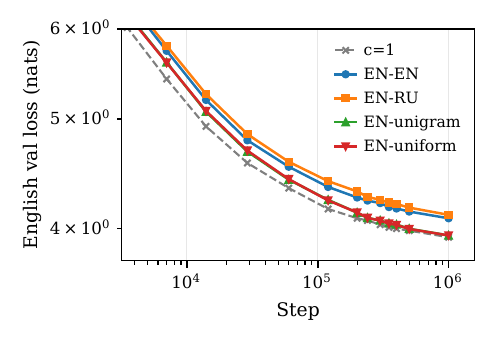}
    \caption{English val loss over training.}\label{fig:cap-val}
  \end{subfigure}
  \hfill
  \begin{subfigure}[b]{0.49\textwidth}
    \includegraphics[width=\linewidth]{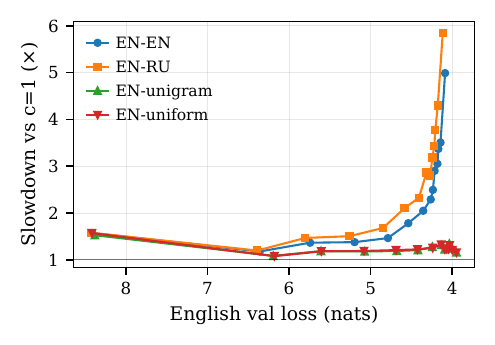}
    \caption{Slowdown vs.\ the single-compartment baseline.}\label{fig:cap-slow}
  \end{subfigure}
  \caption{\textbf{The amount of structure in each compartment determines the capacity cost of compartmentalization.} All runs use a 14.7M base model with $c{=}2$; we compare four choices for the second compartment's data---English (a homogeneous control), Russian, unigram-frequency-sampled noise, and uniform-random-token noise---against the single-compartment ($c{=}1$) baseline.
  \subref{fig:cap-val} English-side validation loss for each condition.
  \subref{fig:cap-slow} Slowdown of each $c{=}2$ run vs.\ the $c{=}1$ baseline at matched English val loss. When both compartments contain language data, the English sample efficiency plateaus vs. baseline, but with noise, English data matches baseline sample efficiency.}
  \label{fig:capacity-sharing}
\end{figure}

To systematically explore compartmentalization, we first contribute a simple method of creating $c$ distinct presentations of statistically identical data. Because data from these compartments is statistically identical, in the best case, models could theoretically share a tremendous amount of representational and computational structure across them, but in the worst case, they would waste modeling capacity by treating them as $c$ independent tasks.

To study compartmentalization in isolation from the structural cues that encourage representation sharing in natural data (subword overlap, syntactic regularity, semantic correspondence), \textbf{we create a maximally compartmentalized model with $c$ presentations with a tokenization trick\footnote{\citet{schafer2024cloned} use an equivalent cloned-vocabulary construction to study cross-lingual generalization under language imbalance.}: increasing the base size of the tokenizer vocabulary $V$ by $c$ to obtain a tokenizer with $cV$ tokens. Then, to encode data in presentation $j$, we simply offset token IDs by $jV$}. Although this allows us to encode arbitrary ratios of data in each presentation, we assume an even $c$-way split in our experiments. In this construction, the effects of compartmentalization are equally as bad for each compartment. We initialize the larger joint tokenizer in the normal way, without any a priori shared structure.

We address both sample efficiency and capacity by training multiple scales of small models to convergence on a fixed dataset with $\sim$6--5000x more tokens than is Chinchilla-optimal \citep{hoffmann2022chinchilla} at each scale. Specifically, we train GPT-style decoders (\S\ref{sec:appendix-arch-model}) on $\sim$131 billion tokens of FineWeb \citep{penedo2024fineweb}, i.e., 1M steps with a batch size of 2048, a block size of 64, and model sizes (see Table \ref{tab:model-sizes}) we can train on a single GPU due to compute constraints. We train all experiments with a learning rate of 2e-5 to avoid instability while overtraining, and no weight decay, which we found had no effect on fully-compartmentalized models\footnote{We additionally swept weight decay over $\{0, 0.01, 0.05, 0.1, 0.2\}$ at the 14.7M scale across $c \in \{2, \ldots, 8\}$; final val-loss spread was $\leq 0.03$ nats}. We accept the additional limitation of using a 16k tokenizer we train ourselves, breaking any perplexity comparisons with established models, to mitigate the memory cost of multiplying the vocabulary size up to eight times.

The results of our experiments are in Figure \ref{fig:efficiency_capacity}.
The slowdown is defined as the ratio describing the multiple of steps required to achieve a given validation loss for a compartmentalized model compared to the baseline $c{=}1$ model. \textbf{Supporting our characterization of compartmentalization, we observe \textit{sample efficiency} slowdown vs. $c{=}1$ increasing with $c$, and validation loss plateauing higher with $c$, indicative of \textit{capacity competition}}.
Notably, we train fewer models at the $\sim$1B scale due to compute constraints.

We next present two findings related to capacity: \textbf{(a)} The capacity cost we observe in compartmentalized models is not fundamental, and unified solutions exist. \textbf{(b)} Models internally allocate capacity based on the amount of structure within each compartment.

\subsection{Existence of unified solutions}
\label{sec:existence}

\begin{figure}[t]
  \centering
  \begin{subfigure}[b]{0.32\textwidth}
    \includegraphics[width=\linewidth]{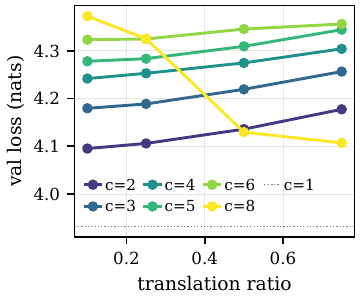}
    \caption{Final val loss vs translation ratio.}\label{fig:tr-nowd-val}
  \end{subfigure}
  \hfill
  \begin{subfigure}[b]{0.32\textwidth}
    \includegraphics[width=\linewidth]{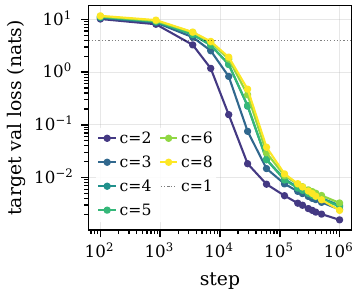}
    \caption{Translation-target loss ($\text{tr}{=}0.5$).}\label{fig:tr-nowd-target}
  \end{subfigure}
  \hfill
  \begin{subfigure}[b]{0.32\textwidth}
    \includegraphics[width=\linewidth]{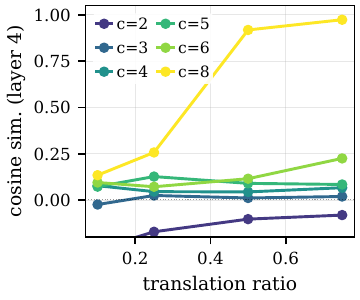}
    \caption{Cross-compartment cossim.}\label{fig:tr-nowd-cossim}
  \end{subfigure}
  \caption{\textbf{For the 14.7M param configuration, parallel ``translation'' data has a slightly negative effect until a phase transition at $c{=}8$. The translation task itself is solved at every $c$, but does not serve to reduce compartmentalization.}
  \subref{fig:tr-nowd-val} Final per-compartment validation loss vs.\ translation ratio, one line per compartment count $c$. Dotted line marks the $c{=}1$ baseline floor (3.932 nats).
  \subref{fig:tr-nowd-target} Target-half validation loss is the cross-entropy on the second half of paired translation sequences---predicting compartment-$j$ tokens given a compartment-$i$ source, averaged across all $i \neq j$ pairs.
  Every line falls below 0.04 nats by step ${\sim}60$k and continues to memorize toward ${\sim}0.002$ nats by the end of training.
  \subref{fig:tr-nowd-cossim} Mean per-token cosine similarity between middle-layer activations of the same canonical text encoded in different compartments, vs.\ translation ratio.}
  \label{fig:tr-no-wd}
\end{figure}

Results in the previous section raise a natural question: is it even possible for our models to share strength across compartments? The answer is yes:
we demonstrate that there exist solutions for compartmentalized prediction with sample efficiency matching the baseline in two complementary ways: initialization duplication and post-hoc parameter duplication.
In other words, while it is \emph{possible} for LLMs to share strength across compartments, they regularly fail to do so.

\subsubsection{Initialization duplication}

In this condition, instead of initializing all tokens independently, we initialize only the first $V$ tokens corresponding to the base vocabulary, and copy their values to every other range $Vj\dots V(j + 1)$, without aliasing the ranges to each other. For example, at initialization, the embedding for ``apple'' is identical across all compartments, but these are allowed to diverge during training.

\paragraph{Results} An initialization-copied model matches baseline sample efficiency, exhibiting no relative capacity plateau. We read this as an existence proof: a parameter setting where compartmentalization is ``free'' exists and is reachable by SGD, but not from a generic initialization. Refer to the appendix, Figure~\ref{fig:copyemb-slowdown}, for the slowdown comparison.

\subsubsection{Post-hoc parameter duplication}

We additionally find that it is possible to convert a $c{=}1$ model into a $c\in\{2,4,8\}$ model without the capacity overhead of training it from scratch by duplicating the learned word embeddings and corresponding LM head entries after 1M steps. For all $c$, only 2\% of initial 1M steps are required to overcome an initial loss spike, after which the model continues on the original plateau (Fig.~\ref{fig:copyemb-posthoc}).

\subsection{Capacity sharing behavior}

\begin{figure}
  \centering
  \begin{subfigure}[b]{0.49\textwidth}
    \includegraphics[width=\linewidth]{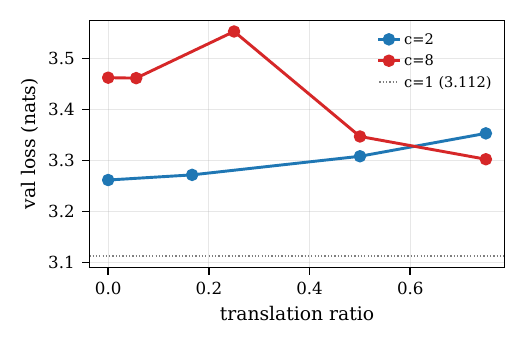}
    \caption{Final validation loss at $\sim$1B.}\label{fig:1b-val}
  \end{subfigure}
  \hfill
  \begin{subfigure}[b]{0.49\textwidth}
    \includegraphics[width=\linewidth]{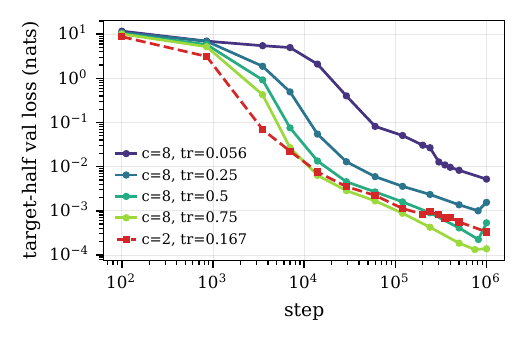}
    \caption{$\sim$1B: Target-half validation loss vs.\ training step.}\label{fig:1b-target}
  \end{subfigure}
  \caption{\textbf{The $c{=}8$ phase transition replicates at 1B parameters between $\text{tr}{=}0.25$ and $\text{tr}{=}0.5$, and the local translation task is mastered increasingly faster as the ratio grows.}
  \subref{fig:1b-val} Final per-compartment validation loss as a function of translation ratio, matched at step $10^{6}$. $c{=}2$ increases from $+0.15$ to $+0.24$ as the ratio increases. $c{=}8$ stays at the $c{=}8$ plateau ($+0.35$--$0.44$ nats) for $\text{tr} \le 0.25$ and breaks to $+0.19$--$+0.24$ nats for $\text{tr}\ge 0.5$.
  \subref{fig:1b-target} Cross-entropy on the second (target) half of paired translation sequences vs.\ training step. Even the lowest ratio ($0.056$) reaches ${\sim}5\times10^{-3}$ nats by training end.}
  \label{fig:1b}
\end{figure}

In Figure \ref{fig:capacity-sharing}, we provide further evidence that the sample efficiency plateaus we observe are, in fact, the result of tasks being forced to share a finite amount of model capacity. Additionally, this reinforces that data do not consume representational capacity proportional to frequency or relative loss alone; instead, capacity appears to be allocated to the data with the most recognizable structure.

\section{Interventions for compartmentalization}

We have shown that there are parameter settings that share statistical strength across compartments, but that state-of-the-art training algorithms fail to find them.
How might we adjust architectures, initializations, or algorithms to maximize the sharing of structure?

We discuss two approaches to identify and reduce compartmentalization in our simplified setting, both of which assume varying amounts of parallel data are available.

\subsection{Paired data is often insufficient}

We demonstrate that small language models can quickly learn a simple mapping between presentations but fail to use it to improve compartmentalization. Taking inspiration from multilingual learning, we encode an experimentally controlled ratio of the dataset as ``translation data,'' where the same text is presented twice, with each half encoded in one of two randomly selected compartments, and translation task tokens preceding each presentation.

Additionally, to avoid requiring the model to spend additional capacity learning $c$-way compartment identity for common tokens in each compartment, we introduce compartment embeddings, i.e., a $d_{\text{model}}$ embedding vector for each of $c$ compartments that we add to the word embedding for each token encoded within that compartment. This includes translation examples, where we add distinct compartment embeddings to each half, including their preceding translation task tokens. In particular, this removes any ambiguity from the second half of each example, making it completely predictable and allowing us to use loss convergence toward zero on held-out examples to determine whether the translation task is learned.

\paragraph{Results} See Figure~\ref{fig:tr-no-wd}. For our 14.7M--73.4M parameter configuration, we sweep $c$ and translation ratio, but hold weight decay at 0---we investigate the effect of weight decay in the next section. Surprisingly, we observe that for most $c$, translation data have almost no effect on decreasing final validation loss, i.e., compartments compete for capacity despite the model quickly learning the translation task (Fig.~\ref{fig:tr-nowd-target}). Even more surprisingly, we observe that under some conditions---here, at $c{=}8$---the model is capable of using translation data to break the compartmentalization plateau.

We perform limited experiments at $\sim$1B scale, due to compute constraints, which sharpen these observations at a larger scale: very little translation data (at $c{=}2$, $tr{=}0.17$, and at $c{=}8$, $tr{=}0.06$) are needed for the model to learn the translation task to convergence, but for $c{=}8$, the validation loss plateau does not break until $tr{=}0.5$, which represents a costly improvement considering that half of the data are encoded as translation pairs.

\textbf{Taken together, we propose this as an existence proof that language models can struggle to recognize opportunities in data to create unified representations, even well into convergence.}

We investigate the effects of weight decay on our 14.7M parameter model next.

\subsubsection{Weight decay accelerates, reinforces the paired-data phase transition}

\begin{figure}
  \centering
  \begin{subfigure}[b]{0.49\textwidth}
    \includegraphics[width=\linewidth]{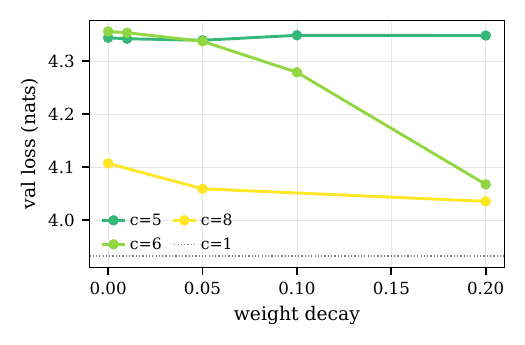}
    \caption{Final per-compartment validation loss.}\label{fig:wd-tr075-val}
  \end{subfigure}
  \hfill
  \begin{subfigure}[b]{0.49\textwidth}
    \includegraphics[width=\linewidth]{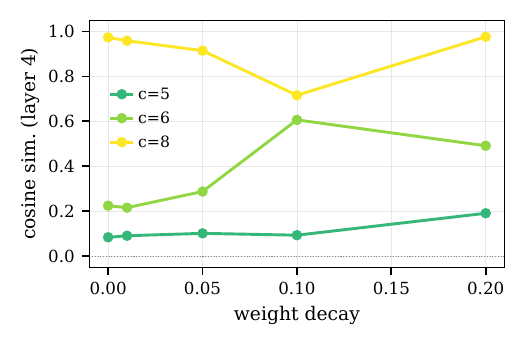}
    \caption{Cross-compartment cosine similarity (layer 4).}\label{fig:wd-tr075-cossim}
  \end{subfigure}
  \caption{\textbf{At translation ratio 0.75, weight decay shifts the validation plateau phase transition from $c{=}8$ down to $c{=}6$.}
  14.7M base model, $\text{tr}{=}0.75$ (absolute).
  \subref{fig:wd-tr075-val} Validation loss as a function of weight decay. $c{=}5$ is unaffected by weight decay---the line stays at the $c{=}5$ plateau (${\sim}4.34$) for every $\text{wd} \in \{0, 0.01, 0.05, 0.1, 0.2\}$. $c{=}6$ inflects sharply between $\text{wd}{=}0.05$ and $\text{wd}{=}0.2$, dropping to ${\sim}4.07$---within $0.03$ nats of the best $c{=}8$ cell anywhere in our sweep ($4.035$ at $\text{wd}{=}0.2$).
  \subref{fig:wd-tr075-cossim} Mean per-token cosine similarity between layer-4 activations of the same canonical text encoded under different compartments, averaged over compartment pairs. Despite the matched validation loss at $\text{wd}{=}0.2$, there is only an inconclusive weight-decay-driven trend in cosine similarity at $c{=}6$.
  }
  \label{fig:wd-tr075}
\end{figure}

In Figure~\ref{fig:wd-tr075}, we report the results of additional experiments at the 14.7M scale in which we swept the weight decay hyperparameter $\lambda \in \{0, 0.01, 0.05, 0.1, 0.2\}$. The most striking result is an acceleration of the phase transition, which happens earlier, at $c{=}6, tr{=}0.75$, as $\lambda$ increases. Additionally, the existing $c{=}8$ transition is reinforced and accelerated, with $c{=}8, tr{=}0.25$ at 4.035 nats, lower than any loss achieved at $c{=}2$. \textbf{This provides some preliminary evidence that aggressive weight decay may help language models learn from data to reduce compartmentalization.}

\subsection{Contrastive representation alignment}

\begin{figure}[t]
  \centering
  \begin{subfigure}{0.49\textwidth}
    \centering
    \includegraphics[width=\linewidth]{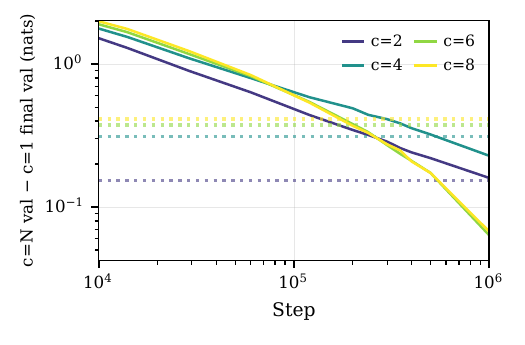}
    \caption{Residual val loss to $c{=}1$ floor.}
    \label{fig:infonce-c-dep-loss}
  \end{subfigure}
  \hfill
  \begin{subfigure}{0.49\textwidth}
    \centering
    \includegraphics[width=\linewidth]{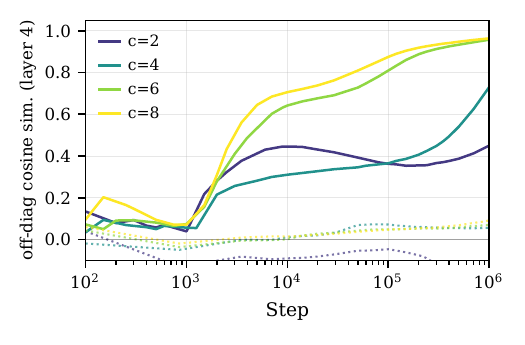}
    \caption{Cross-compartment cosine sim.\ (layer 4).}
    \label{fig:infonce-c-dep-cossim}
  \end{subfigure}
  \caption{\textbf{InfoNCE's benefit scales with $\mathbf c$: at $\mathbf{c{=}6}$ and $\mathbf{c{=}8}$ it recovers $\mathbf{\sim83\%}$--$\mathbf{\sim84\%}$ of the compartmentalization tax, at $\mathbf{c{=}4}$ it closes $\mathbf{\sim26\%}$, and at $\mathbf{c{=}2}$ it provides no measurable improvement.}
  \subref{fig:infonce-c-dep-loss} Residual val loss of each InfoNCE-trained $c$-compartment model relative to the fully-trained $c{=}1$ baseline (3.932 nats). At $c{\in}\{6,8\}$, InfoNCE breaks each respective plateau baseline and approaches $c{=}1$ (residual $+0.063$ and $+0.067$ nats vs.\ $c=1$ floor $3.932$; $\sim83\%$ and $\sim84\%$ of the gap closed, respectively). At $c{=}4$, InfoNCE closes $\sim26\%$ of the gap ($+0.228$ vs. $+0.310$ without InfoNCE). At $c=2$, we find no measurable benefit ($+0.160$ vs. $+0.154$).
  \subref{fig:infonce-c-dep-cossim} Mean off-diagonal cosine similarity between layer 4 paired-token activations across compartments. Cosine similarity increases most dramatically after the phase transition, i.e., for $c{=}6$ and $c{=}8$.}
  \label{fig:infonce-c-dep}
\end{figure}

In addition to dataset augmentations, we explore a training objective modification, using an auxiliary contrastive objective, InfoNCE \citep{oord2019infonce}, to directly align mean-pooled middle layer representations for the same text encoded in two different compartments. We use an overall batch size of 32 for InfoNCE, i.e., one positive example with 31 negatives per step, and we found by sweeping $\lambda$ coefficient we applied to the auxiliary loss across $\{0.1, 0.7, 1.0, 1.3, 10\}$ for $c{=}2$ that $\lambda \in [0.7, 1.3]$ performed within 0.013 nats of each other; we use $\lambda{=}1.0$ throughout.

\textbf{Our results (Fig.~\ref{fig:infonce-c-dep}) replicate a phase transition similar to what we observe with paired data}, but our best conditions, $c{=}8$ and $c{=}6$, close the gap to $c{=}1$ more effectively than any of our paired data conditions. This is also very costly, given that our InfoNCE batches sample the entire dataset with replacement, but more concerning is the fact that such an extreme intervention is insufficient for $c{=}2$ and only marginally beneficial for $c{=}4$. We do not have a definitive mechanistic explanation for this asymmetric behavior and leave a careful study to future work.

\section{Case studies of compartmentalization}
\label{sec:case-studies}

As an additional contribution, we propose tokenizer compartmentalization (\S\ref{sec:worst-case}) as a baseline for measuring representation sharing between two or more identifiable presentations in natural data---multiple languages, multiple programming languages, or multiple paraphrases of the same fact. The construction compares three matched training runs: (i) a model trained on a single presentation, (ii) a model trained on all presentations with a shared tokenizer (the standard setting), and (iii) a model trained on the same presentations but with a disjoint tokenizer per presentation, as in \S\ref{sec:worst-case}.

Conditions (i) and (iii) anchor the measurement. (i) is a per-presentation ceiling when all capacity is devoted to one presentation, and (iii) is a no-sharing, no-interference baseline in which each presentation receives its own parameter slabs. The position of (ii) characterizes joint training relative to both. Performance close to (i) indicates strong cross-presentation sharing; performance close to (iii) indicates joint training has failed to amortize across presentations; performance \textit{worse than} (iii) indicates destructive capacity competition---presentations interfering in shared parameters in ways isolation would prevent.

Our two case studies cover different regimes. Multilingual joint training (\S\ref{sec:case-studies}.1) at the scales we study falls between (i) and (iii), indicating partial sharing. Joint biography/Q\&A training (\S\ref{sec:bio-competition}) falls below (iii), indicating destructive competition. Both assume matched architecture, hyperparameters, and per-step batch composition across the three conditions.

\subsection{Sample inefficiency: measuring bilingual compartmentalization}
\label{sec:bilingual}

\begin{figure}[t]
  \centering
  \begin{subfigure}[b]{0.49\textwidth}
    \includegraphics[width=\linewidth]{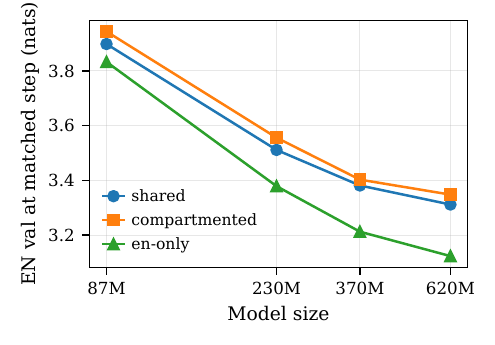}
    \caption{Matched-step EN val loss.}\label{fig:ml-en}
  \end{subfigure}
  \hfill
  \begin{subfigure}[b]{0.49\textwidth}
    \includegraphics[width=\linewidth]{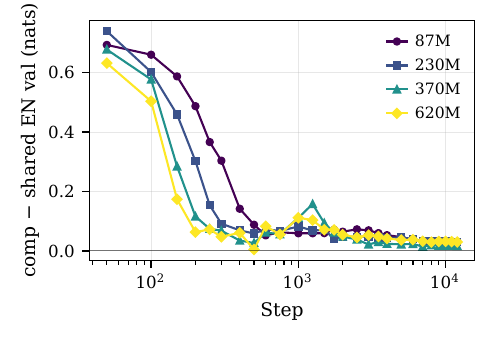}
    \caption{Compartmentalized vs. shared val loss.}\label{fig:ml-gap}
  \end{subfigure}
  \caption{\textbf{At <1B scale, the gap between multilingual and compartmentalized-multilingual validation loss is small, but there is a consistent sample-efficiency gap between English-only and multilingual.}
  \subref{fig:ml-en} EN validation loss at step 5000 (the final step of the en-only run) for shared, compartmentalized, and en-only models, across four scales (87M to 620M parameters).
  \subref{fig:ml-gap} The validation loss gap between training with a single multilingual tokenizer vs. two completely disjoint tokenizers (compartmentalized). The gap between the two closes quickly with steps, suggesting that at least early in training, and for small models, there is high multilingual compartmentalization.}
  \label{fig:multilingual-scaling}
\end{figure}

Pretraining 87M parameter Llama-style models using the multilingual Qwen3 \citep{yang2025qwen3} tokenizer on 767M tokens each of English and Chinese Wikipedia text, we find preliminary evidence (Figure \ref{fig:multilingual-scaling}) from our notion of speedup that language models compartmentalize languages at smaller scales and suffer a capacity tax vs. single-language training---notice in particular that the shared curve is much closer to the tokenizer-compartmentalized curve, and that this gap decreases with steps, suggestive, given our sample efficiency observations, of low cross-presentation information sharing. We lacked the access to compute to attempt compute-optimally pretraining larger models across these conditions, so we do not claim that this trend continues with further scale.

Findings from \citet{ifergan2024beneathsurfaceconsistency} and \citet{goldman-etal-2025-eclektic} provide some evidence that models trained on multiple languages could be compartmentalized, at $\sim$7B and frontier scales, respectively, but these experiments are insufficient in scope to confirm this hypothesis. The sample efficiency gap between shared and compartmentalized models could widen with scale, and correspondingly the gap between English and multilingual training could close. We present these findings as a validation of the basic characterization of compartmentalization: training during a period in which the model does not recognize consistencies across presentations.

\subsection{Capacity competition: synthetic biographies and Q\&A pairs}
\label{sec:bio-competition}

We construct a fully synthetic task, inspired by \citet{allenzhu2024physicslanguagemodels31}, where simple attribute-value pair ``facts'' about generated individuals are presented in two forms: as a simple Wikipedia-like biographical format containing all facts about an individual or as a single question-answer pair. The synthetic setting allows us to vary the number of individuals to fill model capacity. We then compare with a new baseline where each of the two fact-presentation formats is tokenized separately to observe capacity competition effects.

We train 35M parameter GPTs (Appendix~\ref{sec:appendix-bio}), otherwise following the architecture introduced in \S\ref{sec:worst-case}. We tuned the size of this synthetic world to resist simple memorization by a model at this scale, settling on 15k synthetic profiles with 10 paraphrases per attribute type in both biography and Q\&A formats. We train for many epochs to encourage memorization; for example, we trained the models for our biography-only condition for 45,000 iterations, where each fact was seen 5,780 times across different paraphrases, and achieved an ultimate fact-recall accuracy of 83.8\% (averaged across three seeds).

\begin{figure}[t]
  \centering
  \begin{subfigure}[b]{0.49\textwidth}
    \includegraphics[width=\linewidth]{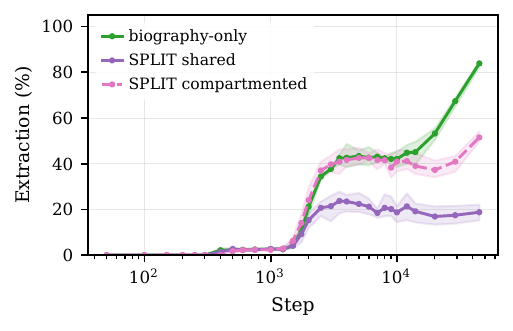}
    \caption{Train: bio. format. Probe: bio. continuation.}
    \label{fig:bio-bio-bio}
  \end{subfigure}
  \hfill
  \begin{subfigure}[b]{0.49\textwidth}
    \includegraphics[width=\linewidth]{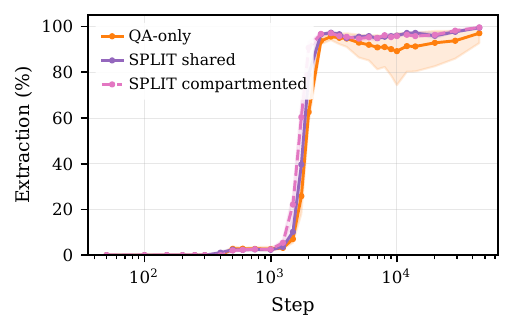}
    \caption{Train: QA format. Probe: QA prompt.}
    \label{fig:bio-qa-qa}
  \end{subfigure}
  \caption{\textbf{Biography vs. Q\&A capacity (N=15k people, seed-averaged): Same data in different formats compete for representational capacity.}
  Each panel: extraction accuracy vs.\ training step (log) for four training conditions, mean across 3 seeds.
  We show only same-format results here; cross-format results (Fig.~\ref{fig:bio-appendix}) only validate that compartmentalization completely prevents information sharing across formats, with off-diagonal accuracies of $\sim$0\%.}
  \label{fig:bio-capacity-seedavg}
\end{figure}

\paragraph{Capacity effects} In Fig.~\ref{fig:bio-bio-bio}, we compare accuracy across training for three training conditions, averaged across seeds: only biography data, both biography and Q\&A data, and the two formats tokenizer-compartmentalized. The compartmentalized baseline reveals a capacity effect: the highest score is the 83.8\% we discussed, followed by a sharp drop to 51.5\% for the tokenizer-compartmentalized model, and 18.8\% for the baseline two-format model. The fact that in each case the model achieves nearly 100\% accuracy on Q\&A data (Fig.~\ref{fig:bio-qa-qa}) reinforces that when the two formats compete for limited capacity, the less ambiguous one is allocated more. A higher biography accuracy score for the compartmentalized model suggests that compartmentalization frustrates destructive cross-format capacity competition.

\section{Discussion}

Under the worst-case construction (\S\ref{sec:worst-case}), language models pay capacity and sample-efficiency costs that grow with the number of compartments $c$. These costs are not architectural: copy-initialized and post-hoc-duplicated models reach the single-compartment plateau (\S\ref{sec:existence}), so a parameter setting in which compartmentalization is ``free'' exists and is reachable under some initializations, but SGD does not find it well into convergence.

At the scales we study, three interventions partially break the stuck regime, with a consistent pattern: aggressive weight decay sharpens the existing $c{=}8$ phase transition and shifts it down to $c{=}6$ at high translation ratios; paired translation data, learned by the model to near-zero loss in every condition, only reduces compartmentalization once a $c$-dependent threshold is crossed; and an explicit contrastive alignment loss similarly recovers more of the cost as $c$ increases. Without identifying a unifying cause for this trend, we offer this as evidence that language models may be inconsistent in their ability to unify internal representations across disjoint presentations.

Our case studies (\S\ref{sec:case-studies}) suggest the same framework describes failures of small real models. Multilingual joint training closely tracks the disjoint-tokenizer baselines across scale, and a synthetic biography/Q\&A dataset induces capacity competition. We offer compartmentalization as one lens on a class of representational failures the field has documented at scale.

\section{Limitations}

\begin{tightlist}
\item We train small models, with limited experiments at slightly larger scales. It is possible that compartmentalization disappears with scale, despite the evidence we cite. Further, we define only a simple notion of compartmentalization. It could be too inflexible to usefully describe phenomena the community cares about. In addition, our multilingual case study is increasingly undertrained as scale increases, so we make no claims about multilingual compartmentalization in larger models.
\item All single-seed runs in this paper are reported without confidence intervals; the synthetic biography panels (\S\ref{sec:bio-competition}) are the exception, reporting mean across 3 random seeds with shaded bands showing the min/max range across those seeds. We argue robustness through monotonicity across sweep grids and qualitative replication of headline phenomena across multiple model scales (Fig.~\ref{fig:efficiency_capacity}\subref{fig:plateau}); due to compute constraints, we do not test InfoNCE at scales other than 14.7M.
\end{tightlist}

See Appendix~\ref{sec:appendix-extended-limitations} for extended discussion on the limitations of compartmentalization.


\bibliographystyle{plainnat}
\bibliography{references}

\appendix


\section{Architecture and training setup}
\label{sec:appendix-arch}

\subsection{Model architecture}
\label{sec:appendix-arch-model}

\begin{table}[ht]
\centering
\caption{Model sizes for the compartment baseline experiments. The composite vocab in the $n$-compartment setting is $nV + 1$ (the $+1$ is the cross-compartment translation token). ``Trunk'' counts attention and MLP parameters only.}
\label{tab:model-sizes}
\
\begin{tabular}{lccrrr}
\toprule
Config & $L$ & $d_{\mathrm{model}}$ & Trunk & Total ($c{=}1$) & Total ($c{=}8$) \\
\midrule
8-32           & 8  & 32   & \phantom{00}0.1\,M & \phantom{00}1.2\,M & \phantom{00}8.5\,M \\
8-64           & 8  & 64   & \phantom{00}0.4\,M & \phantom{00}2.5\,M & \phantom{0}17.2\,M \\
8-128          & 8  & 128  & \phantom{00}1.6\,M & \phantom{00}5.8\,M & \phantom{0}35.1\,M \\
8-256          & 8  & 256  & \phantom{00}6.3\,M & \phantom{0}14.7\,M & \phantom{0}73.4\,M \\
8-512          & 8  & 512  & \phantom{0}25.2\,M & \phantom{0}41.9\,M & 159\,M \\
\addlinespace
1B (24-1792)   & 24 & 1792 & 925\,M             & 984\,M             & 1.39\,B \\
\bottomrule
\end{tabular}
\end{table}

All models are GPT-style decoders following the nanoGPT \citep{karpathy2022nanogpt} lineage with two modifications: (a) RoPE \citep{su2023rope} replaces learned absolute position embeddings, and (b) embedding and LM-head matrices are not tied. We use pre-LayerNorm, GELU MLPs with $4d_{\mathrm{model}}$ hidden width, causal attention with $d_{\mathrm{head}}{=}32$ (so $n_{\mathrm{head}} = d_{\mathrm{model}}/32$), block size $T{=}64$, and no dropout. Table~\ref{tab:model-sizes} summarizes parameter counts at $c{=}1$ and $c{=}8$ across the model-size ladder.

\subsection{Optimizer and schedule}
\label{sec:appendix-arch-optim}

We use AdamW with $\beta_1{=}0.9$, $\beta_2{=}0.95$, $\varepsilon{=}10^{-8}$, and a peak learning rate of $2 \times 10^{-5}$ reached via $1{,}000$ iterations of linear warmup (small-scale runs) or $1{,}000$--$2{,}000$ iterations of linear warmup (1B runs); learning rate is then held constant for the remainder of training, with no decay. We train for $1{,}000{,}000$ iterations at the small-scale ladder (8-32 through 8-512) and approximately the same number of optimizer steps at the 1B scale, using bfloat16 mixed precision throughout. The effective batch size is held at $2048 \times T = 131{,}072$ tokens per optimizer step across all small-scale runs, with per-step batch size and gradient-accumulation factor varied to fit the VRAM envelope of each $(d_{\mathrm{model}}, c)$ pair on a single 80\,GB GPU. Weight decay is $0$ except in the sweep of \S\ref{sec:appendix-fullgrid-wd}.

\subsection{Compute resources}
\label{sec:appendix-arch-compute}

Experiments ran on a shared cluster of A100-80GB nodes; 1B-scale runs additionally used a B200/A100 instance with 8-GPU single-node DDP. The same effective batch size was used across both types of GPUs. Approximately 2{,}000--2{,}500 A100-equivalent GPU-days were used for the experiments reported in this paper, dominated by the 1B-scale translation-ratio sweep (\textasciitilde 1{,}800 GPU-days). Total compute for the full research project, including exploratory work, is approximately 4{,}000 A100-equivalent GPU-days.

\subsection{Tokenizer}
\label{sec:appendix-arch-tokenizer}

For the worst-case-model and intervention experiments we train a BPE tokenizer of vocabulary size $V{=}16{,}384$ on a subsample of FineWeb \citep{penedo2024fineweb} using SentencePiece's BPE implementation. The compartmented composite vocabulary is $cV + 1$, where the $+1$ is a single shared cross-compartment translation marker that we prepend to translation-paired sequences. For the multilingual case study we instead use the Qwen3 \citep{yang2025qwen3} tokenizer ($V{\approx}151{,}000$).

\section{Compartmentation construction}
\label{sec:appendix-compartment}

\subsection{Vocabulary offset scheme}
\label{sec:appendix-compartment-offset}

Given a base BPE tokenizer of vocabulary size $V$, the $c$-compartment composite vocabulary is constructed by replicating the base vocabulary $c$ times with offset: token id $t$ encoded under compartment $j$ becomes $jV + t$, where $j \in \{0, \ldots, c{-}1\}$. Embedding and LM-head rows are initialized independently across compartments, so at $t{=}0$ the model has no information that the $c$ slabs of the embedding matrix represent the same underlying tokens.

\subsection{Compartment embeddings}
\label{sec:appendix-compartment-embed}

To prevent the model from spending capacity to learn compartment identity from token co-occurrence, we add a learned $d_{\mathrm{model}}$-dimensional bias to the input embedding of every token according to its compartment, in addition to the offset described above. Compartment embeddings are initialized to zero and applied uniformly within each compartment, including translation-paired examples (each half receives the bias of its own compartment). This makes the second half of each translation pair fully predictable in principle, allowing us to use convergence of the second-half loss toward zero as a check that the translation task is being learned.

\section{Datasets}
\label{sec:appendix-data}

\subsection{FineWeb pretraining corpus}
\label{sec:appendix-data-fineweb}

For the worst-case-model and intervention experiments we use a deduplicated subsample of FineWeb \citep{penedo2024fineweb}, tokenized with the BPE-16384 tokenizer described above. Train and validation shards are disjoint document-level. Across all small-scale runs, training consumes approximately $131{,}072 \times 10^{6} \approx 131$ billion tokens---between $\sim$6$\times$ and $\sim$5000$\times$ Chinchilla-optimal for the model sizes in Table~\ref{tab:model-sizes}.

\subsection{Synthetic biographies and Q\&A pairs}
\label{sec:appendix-data-bio}

The bio capacity dataset (\S\ref{sec:appendix-bio}) consists of $N{=}15{,}000$ synthetic individuals with $A$ structured attributes each (birth date, birthplace, employer, etc.). For each $(\text{person}, \text{attribute})$ pair we generate $K{=}10$ phrasings using a small fixed template bank, yielding $15{,}000 \times A \times 10$ statements in each of two formats: a Wikipedia-style biographical paragraph (BIO) and a single-attribute question-answer pair (QA). The two formats present identical underlying facts; capacity competition between them is the experimental signal.

\subsection{Multilingual EN/ZH Wikipedia}
\label{sec:appendix-data-multi}

For the multilingual case study we use a 767M-token English subsample and a 767M-token Mandarin Chinese subsample of Wikipedia, both tokenized with the Qwen3 tokenizer. The shared-tokenizer condition uses a single Qwen3 vocabulary; the compartmented condition splits the same data across two disjoint vocabularies (each language tokenized independently and offset, as in \S\ref{sec:appendix-compartment-offset}). Models are Llama-style decoders trained at four scales---$(n_{\mathrm{layer}}, d_{\mathrm{model}}) \in \{(12, 256), (24, 512), (24, 768), (24, 1024)\}$, approximately 87M, 230M, 460M, and 620M parameters respectively---with block size 512, trained for 5000 iterations.

\section{Evaluation methodology}
\label{sec:appendix-eval}

\subsection{Per-compartment validation loss}
\label{sec:appendix-eval-percomp}

At each evaluation, we sample a fixed-seeded validation batch and compute cross-entropy at every token position, restricted to positions assigned to a single compartment. The reported \emph{val loss} for a $c$-compartment model is the unweighted mean of the per-compartment losses---equivalent to the loss the model would achieve on a $1/c$-fraction validation set drawn from each compartment in turn. For $c{=}1$ this collapses to the standard FineWeb val loss.

\subsection{Target-half loss for translation pairs}
\label{sec:appendix-eval-target}

On translation-paired evaluation examples, we report the cross-entropy on the \emph{target} half of each pair only (i.e., predicting compartment-$j$ tokens given a compartment-$i$ source for $i \neq j$). This isolates whether the model has learned the translation task itself, separately from whether translation-paired training data has reduced compartmentation in the per-compartment val loss. We average over all $i \neq j$ ordered pairs.

\subsection{Representational alignment}
\label{sec:appendix-eval-cossim}

We measure cross-compartment representational alignment using per-token mean cosine similarity. For a fixed canonical batch of $B{=}64$ FineWeb val token sequences of length $T{=}64$, and for each compartment $j \in \{0, \ldots, c{-}1\}$, we forward $\mathbf{x}_j = \mathbf{x} + jV$ (with compartment-$j$ compartment embeddings) and capture the post-block hidden state at layer $L/2$ (layer 4 for the 8-layer models). For each unordered compartment pair $(i, j)$ we compute the mean per-row cosine similarity between the corresponding $(B \cdot T, d_{\mathrm{model}})$ activation tensors, then average across all $\binom{c}{2}$ pairs to obtain the reported \emph{mean off-diagonal cosine similarity}.

\subsection{Bio extraction probes}
\label{sec:appendix-eval-bio}

For the bio capacity experiments, we evaluate fact recall by prompting each model with templated probes in each format. The BIO probe presents the first $K{-}1$ phrasings of a person's biography and scores extraction of the held-out attribute; the QA probe issues each $(\text{person}, \text{attribute})$ question and scores whether the model's first-token-correct prediction matches the ground-truth answer. Off-diagonal probes (train BIO, probe QA, and vice versa) measure cross-format generalization; diagonal probes measure same-format recall. We report mean accuracy across the test set, averaged over 3 random seeds.

\subsection{Validation batch averaging}
\label{sec:appendix-eval-batch}

All reported validation losses are means over $N_{\mathrm{eval}}{=}100$ independent $B \times T$ batches per checkpoint, sampled from the held-out validation shards. For the multilingual case study (\S\ref{sec:appendix-data-multi}), per-language validation losses use $B{=}4$ sequences of $T{=}512$ tokens averaged over $N{=}200$ contiguous chunks of the held-out per-language val shards, for $409{,}600$ prediction tokens per language per checkpoint---about twice the main pipeline's $204{,}800$.

\section{Full sweep results}
\label{sec:appendix-fullgrid}

\subsection{Weight decay $\times$ translation ratio $\times$ $c$}
\label{sec:appendix-fullgrid-wd}

\begin{figure}[t]
  \centering
  \begin{subfigure}[b]{\textwidth}
    \includegraphics[width=\linewidth]{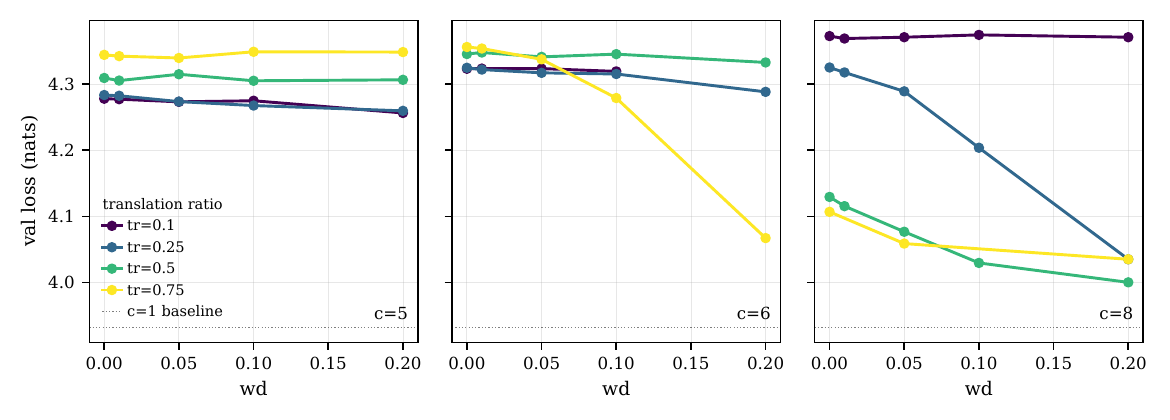}
    \caption{Validation loss.}
    \label{fig:appendix-wd-val}
  \end{subfigure}\\[6pt]
  \begin{subfigure}[b]{\textwidth}
    \includegraphics[width=\linewidth]{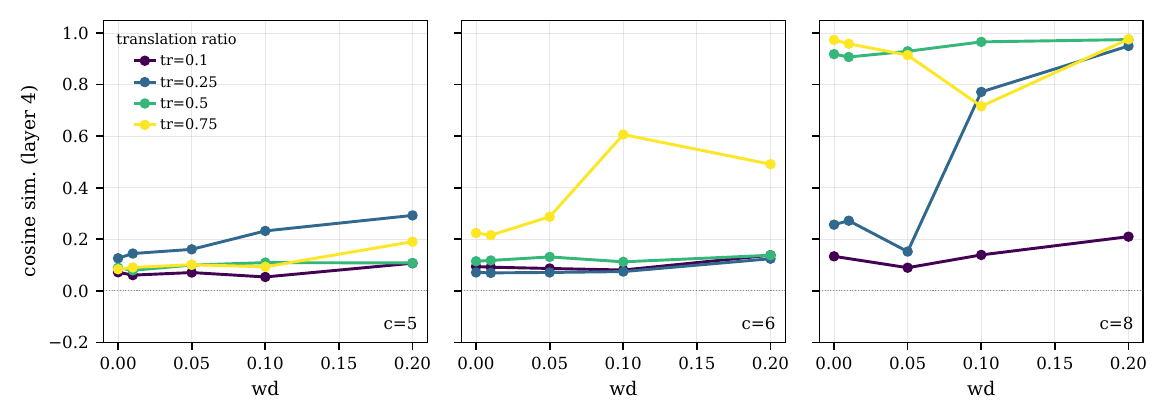}
    \caption{Cross-compartment cosine similarity.}
    \label{fig:appendix-wd-cossim}
  \end{subfigure}
  \caption{\textbf{Full weight-decay sweep at the 14.7M scale, for $c \in \{5, 6, 8\}$ across $\text{tr} \in \{0, 0.056, 0.167, 0.25, 0.5, 0.75\}$ and $\text{wd} \in \{0, 0.01, 0.05, 0.1, 0.2\}$.} Body Fig.~\ref{fig:wd-tr075} shows the $\text{tr}{=}0.75$ slice. $c{=}5$ is weight-decay-immune at every translation ratio; $c{=}6$ inflects only at $\text{tr}{=}0.75$ for $\text{wd}{\geq}0.1$; $c{=}8$ deepens the existing $\text{tr}{\geq}0.5$ phase transition with weight decay.}
  \label{fig:appendix-wd-fullgrid}
\end{figure}

See Fig.~\ref{fig:appendix-wd-fullgrid}.

\subsection{InfoNCE coefficient sweep}
\label{sec:appendix-fullgrid-infonce-lambda}

\begin{figure}[t]
  \centering
  \begin{subfigure}[b]{0.49\textwidth}
    \includegraphics[width=\linewidth]{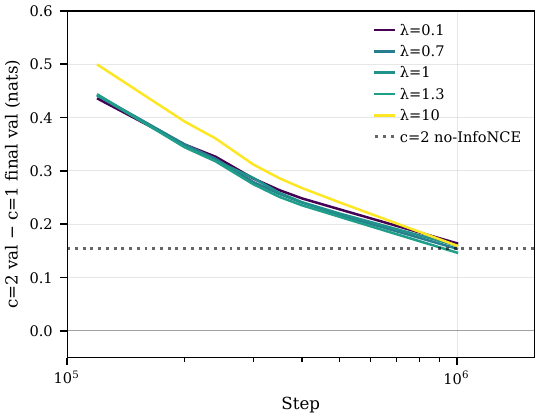}
    \caption{$\lambda$ sweep.}\label{fig:appendix-infonce-lambda}
  \end{subfigure}
  \hfill
  \begin{subfigure}[b]{0.49\textwidth}
    \includegraphics[width=\linewidth]{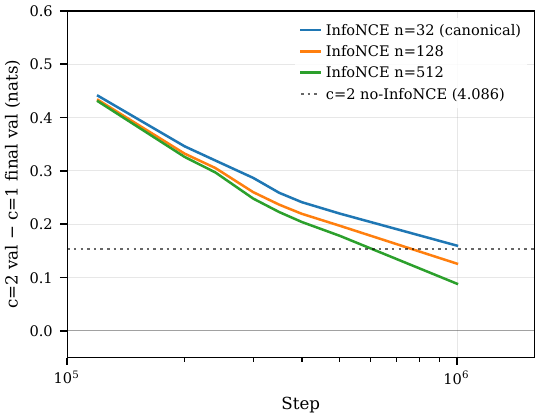}
    \caption{Negative-count sweep at $\lambda{=}1$.}\label{fig:appendix-infonce-batch}
  \end{subfigure}
  \caption{\textbf{InfoNCE hyperparameter tuning at $c{=}2$, 14.7M architecture, $\text{tr}{=}0$.}
  Both panels plot residual val loss relative to the fully-trained $c{=}1$ baseline (3.932 nats); the dotted reference is the $c{=}2$ no-InfoNCE final ($+0.154$ nats).
  \subref{fig:appendix-infonce-lambda} Coefficient sweep $\lambda \in \{0.1, 0.7, 1.0, 1.3, 10\}$ at the canonical $n{=}32$ negatives.
  Step-matched ranking: $\lambda{=}1.3$ ($+0.147$, $4.079$) $<$ $\lambda{=}0.7$ ($+0.154$, $4.086$) $<$ $\lambda{=}1.0$ ($+0.160$, $4.092$) $\approx$ $\lambda{=}10$ ($+0.160$, $4.092$) $<$ $\lambda{=}0.1$ ($+0.164$, $4.096$). The objective is flat across $\lambda \in [0.7, 1.3]$ (within 0.013 nats); we use $\lambda{=}1.0$ throughout.
  \subref{fig:appendix-infonce-batch} Negative-count sweep $n \in \{32, 128, 512\}$ at $\lambda{=}1$. Step-matched values are $n{=}512$ ($+0.088$, $4.020$) $<$ $n{=}128$ ($+0.126$, $4.058$) $<$ $n{=}32$ ($+0.160$, $4.092$). $n{=}512$ closes roughly 43\% of the $c{=}2 \to c{=}1$ gap (vs. 58\% claimed under the old data). No $c{=}2$ run at any setting fully closes the gap to $c{=}1$ within our training budget.}
  \label{fig:appendix-infonce-tuning}
\end{figure}

The body experiments use the canonical setting $\lambda{=}1.0$, $\tau{=}0.1$, sample size $32$ paired sequences per InfoNCE step (one positive plus 31 in-batch negatives per direction, with the loss symmetrized over the two directions), and no cross-rank gather. The alignment loss is applied to mean-pooled hidden states at layer $\lfloor L/2 \rfloor{=}4$. To confirm $\lambda{=}1.0$ was a defensible choice, we swept $\lambda \in \{0.1, 0.7, 1.0, 1.3, 10.0\}$ at $c{=}2$ on the 14.7M architecture, holding all other settings fixed (Fig.~\ref{fig:appendix-infonce-lambda}).

\section{Extended Limitations}
\label{sec:appendix-extended-limitations}

Our notion of compartmentalization is quite simplified, and does not allow for multiple overlapping types of presentations, i.e., under this scoping, surface form must be both discrete and disjoint. We make no claim as to how much compartmentalization as we characterize it contributes to the overall difficulty of modeling language. In addition, while we performed experiments to validate our characterizations of sample efficiency and capacity in our controlled setting, we do not claim that similar behavior in other settings is conclusively or likely caused by representational redundancy.

\section{Bio capacity supplementary}
\label{sec:appendix-bio}

\begin{figure}[t]
  \centering
  \begin{subfigure}[b]{0.49\textwidth}
    \includegraphics[width=\linewidth]{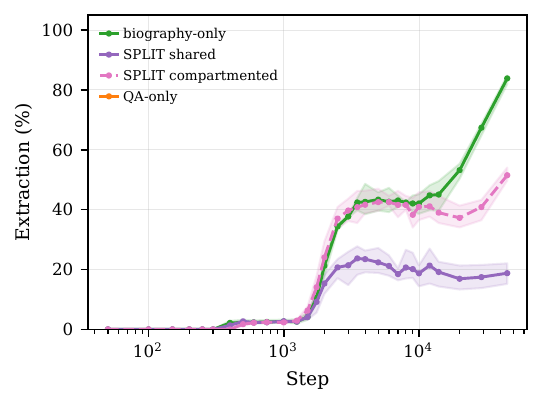}
    \caption{Train: bio.\ Probe: bio.\ continuation.}
    \label{fig:bio-dd}
  \end{subfigure}
  \hfill
  \begin{subfigure}[b]{0.49\textwidth}
    \includegraphics[width=\linewidth]{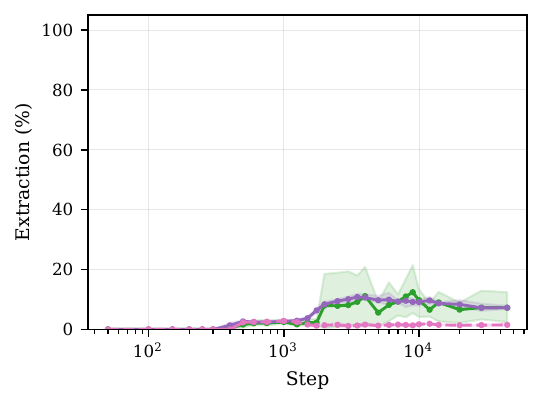}
    \caption{Train: bio.\ Probe: QA prompt.}
    \label{fig:bio-dq}
  \end{subfigure}
  \\[2pt]
  \begin{subfigure}[b]{0.49\textwidth}
    \includegraphics[width=\linewidth]{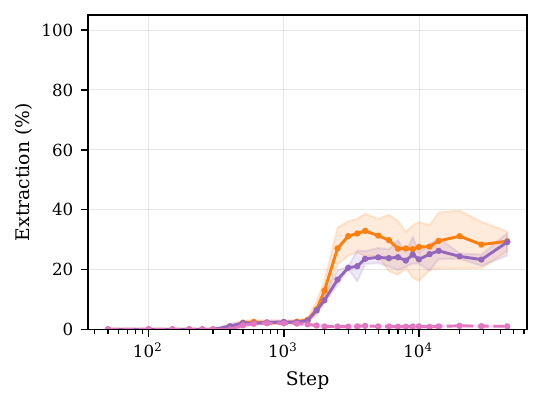}
    \caption{Train: QA.\ Probe: bio.\ continuation.}
    \label{fig:bio-qd}
  \end{subfigure}
  \hfill
  \begin{subfigure}[b]{0.49\textwidth}
    \includegraphics[width=\linewidth]{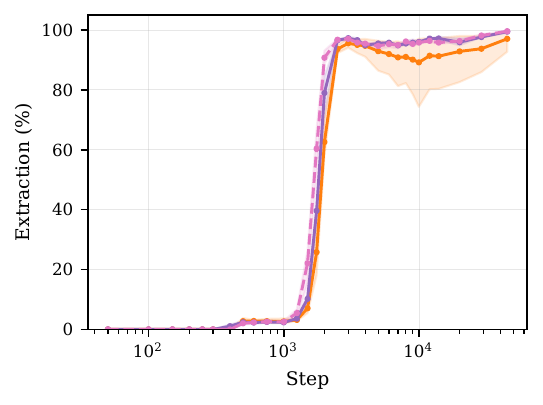}
    \caption{Train: QA.\ Probe: QA prompt.}
    \label{fig:bio-qq}
  \end{subfigure}
  \caption{\textbf{Full extraction-accuracy grid for the bio capacity experiment, $N{=}15{,}000$ profiles, mean across 3 seeds.} Diagonal panels (\subref{fig:bio-dd}, \subref{fig:bio-qq}) are same-format recall; off-diagonal panels (\subref{fig:bio-dq}, \subref{fig:bio-qd}) are cross-format. Compartmented training shows clean format isolation by construction (off-diagonals near $0\%$). The body shows the diagonal panels only (Fig.~\ref{fig:bio-capacity-seedavg}).}
  \label{fig:bio-appendix}
\end{figure}

The bio capacity experiment uses 35M-parameter GPTs with $n_{\mathrm{layer}}{=}12$, $d_{\mathrm{model}}{=}256$, $n_{\mathrm{head}}{=}8$, otherwise matching the architecture in \S\ref{sec:appendix-arch-model}. We train for 45{,}000 iterations on a fixed dataset of $N{=}15{,}000$ synthetic profiles, with each fact seen $\sim$5{,}780 times across distinct paraphrases over the course of training. The compartmented baseline assigns BIO and QA to separate vocab compartments; the joint baseline tokenizes both formats in a single shared vocabulary. We report mean extraction accuracy across 3 random seeds, with the same seed sequence for each condition.

\section{Existence proofs and post-hoc duplication}
\label{sec:appendix-existence}

\subsection{Initialization-copy}
\label{sec:appendix-existence-init}

\begin{figure}[t]
  \centering
  \begin{subfigure}[b]{0.49\textwidth}
    \includegraphics[width=\linewidth]{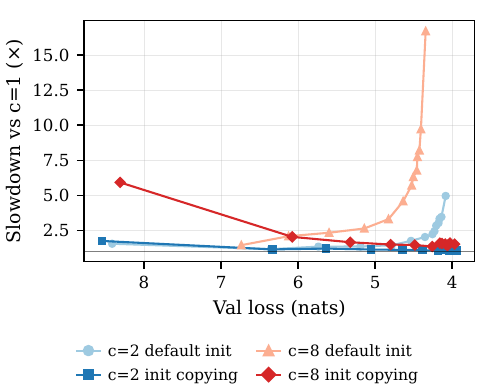}
    \caption{Slowdown for init-copied models vs.\ $c{=}1$ baseline.}\label{fig:copyemb-slowdown}
  \end{subfigure}
  \hfill
  \begin{subfigure}[b]{0.49\textwidth}
    \includegraphics[width=\linewidth]{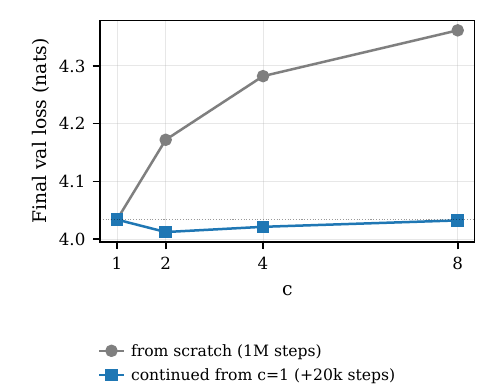}
    \caption{Post-hoc parameter duplication finetuning.}\label{fig:copyemb-posthoc}
  \end{subfigure}
  \caption{\textbf{Validation curves for unified solution experiments.}
  \subref{fig:copyemb-slowdown} Slowdown of compartmented runs vs.\ the $c{=}1$ rope baseline at matched val loss, for $c{=}2$ (blue) and $c{=}8$ (red) at the 14.7M scale. Runs initialised by copying compartment 0's embeddings to every other compartment at $t{=}0$ hug the $c{=}1$ floor throughout training; runs with the standard random init exhibit a capacity-driven sample efficiency plateau.
  \subref{fig:copyemb-posthoc} Minimal continued pretraining (20,000 steps, or 2\% of the steps taken before) on a $c{=}1$ model allows compartment-specific performance matching the paper baseline, i.e. without a capacity plateau.}
  \label{fig:copyemb}
\end{figure}

\begin{figure}[t]
  \centering
  \includegraphics[width=0.7\textwidth]{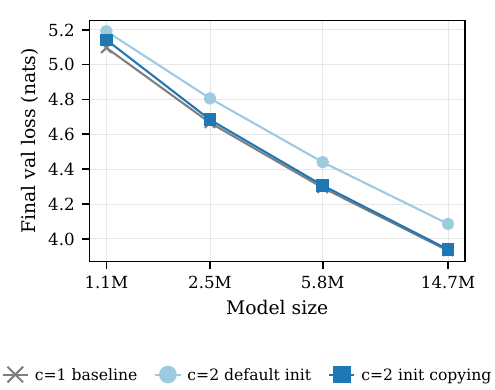}
  \caption{\textbf{Initialization-copy results across $d_{\mathrm{model}} \in \{32, 64, 128, 256\}$.} At every scale, $c{=}2$ with init-copy hugs the $c{=}1$ baseline; default-init $c{=}2$ carries a fixed compartmentation gap. Body Fig.~\ref{fig:copyemb} shows the 14.7M instance only.}
  \label{fig:appendix-copyemb-scaling}
\end{figure}

In Fig.~\ref{fig:copyemb}, we show the training trajectories of embedding copy experiments. In Fig.~\ref{fig:appendix-copyemb-scaling}, we validate that embedding copying works across scales.

\subsection{Post-hoc parameter duplication trajectories}
\label{sec:appendix-existence-finetune}

\begin{figure}[t]
  \centering
  \includegraphics[width=0.7\textwidth]{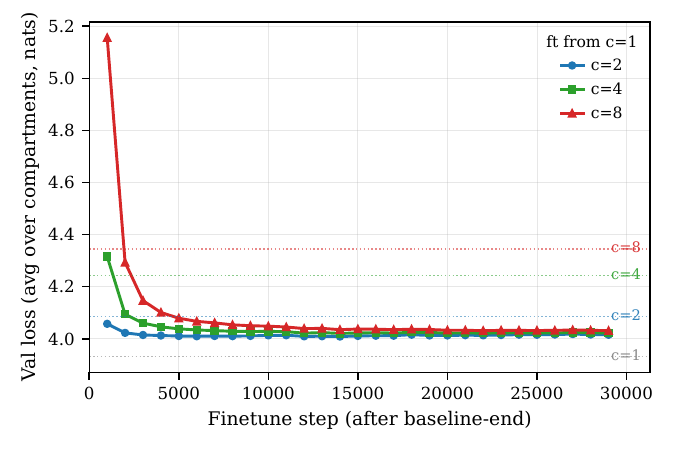}
  \caption{\textbf{Post-hoc parameter duplication, full per-$c$ trajectories at the 14.7M scale.} A $c{=}1$ checkpoint at step $10^6$ is converted into a $c \in \{2, 4, 8\}$ model by duplicating learned word embeddings and LM-head entries. Each curve shows compartment-averaged val loss over the subsequent 20{,}000-step finetune; the initial loss spike is recovered within $\sim$2{,}000 steps and the $c{=}1$ plateau is matched thereafter for every $c$.}
  \label{fig:appendix-finetune-trajectory}
\end{figure}

See Fig.~\ref{fig:appendix-finetune-trajectory}.

\section{Multilingual case study supplementary}
\label{sec:appendix-multi}

\begin{figure}[t]
  \centering
  \begin{subfigure}[b]{0.49\textwidth}
    \includegraphics[width=\linewidth]{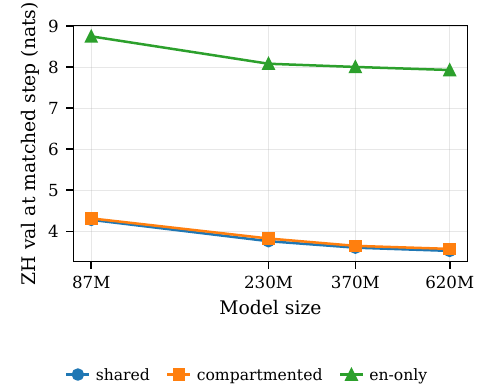}
    \caption{ZH val at matched step.}
    \label{fig:appendix-multi-zh}
  \end{subfigure}
  \hfill
  \begin{subfigure}[b]{0.49\textwidth}
    \includegraphics[width=\linewidth]{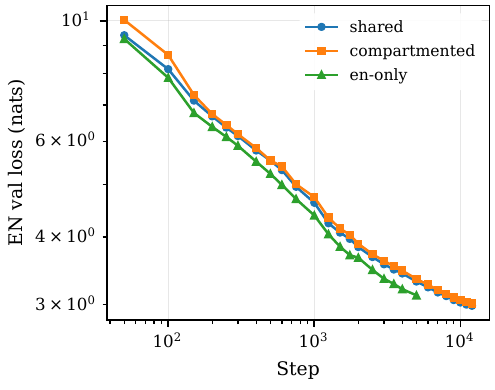}
    \caption{EN val trajectory at 24-1024.}
    \label{fig:appendix-multi-trajectory}
  \end{subfigure}
  \caption{\textbf{Multilingual case study supplementary panels.} Companion to body Fig.~\ref{fig:multilingual-scaling}.
  \subref{fig:appendix-multi-zh} Mandarin-side validation loss at the matched 5000-step horizon, all four scales (counterpart to body Fig.~\ref{fig:multilingual-scaling}\subref{fig:ml-en}).
  \subref{fig:appendix-multi-trajectory} EN val loss over training at the largest multilingual scale (24-1024, 620M parameters), one curve per training condition; the gap between en-only and shared/compartmented is the multilingual capacity tax, visibly closing as training progresses.}
  \label{fig:appendix-multi}
\end{figure}

See Fig.~\ref{fig:appendix-multi}.


\end{document}